\documentclass{article}

\usepackage{microtype}
\usepackage{graphicx}
\usepackage{subfigure}
\usepackage{booktabs} 

\usepackage{epsfig}
\usepackage{graphicx}
\usepackage{amsmath}
\usepackage{amssymb}

\newcommand{\vect}[1]{\boldsymbol{#1}}

\usepackage{multirow}
\usepackage[ruled,algo2e]{algorithm2e}

\usepackage{hyperref}

\usepackage[accepted]{icml2018}

\icmltitlerunning{Gradually Updated Neural Networks for Large-Scale Image Recognition}

\begin{document}

\twocolumn[
\icmltitle{{Gradually Updated Neural Networks for Large-Scale Image Recognition}}



\icmlsetsymbol{equal}{*}

\begin{icmlauthorlist}
\icmlauthor{Siyuan Qiao}{jhu}
\icmlauthor{Zhishuai Zhang}{jhu}
\icmlauthor{Wei Shen}{jhu,su}
\icmlauthor{Bo Wang}{hik}
\icmlauthor{Alan Yuille}{jhu}
\end{icmlauthorlist}

\icmlaffiliation{jhu}{Johns Hopkins University}
\icmlaffiliation{su}{Shanghai University}
\icmlaffiliation{hik}{Hikvision Research}

%
\icmlcorrespondingauthor{Siyuan Qiao}{siyuan.qiao@jhu.edu}

\icmlkeywords{Machine Learning, ICML}

\vskip 0.3in
]



\printAffiliationsAndNotice{}  

\begin{abstract}
Depth is one of the keys that make neural networks succeed in the task of large-scale image recognition.
The state-of-the-art network architectures usually increase the depths by cascading convolutional layers or building blocks.
In this paper, we present an alternative method to increase the depth.
Our method is by introducing computation orderings to the channels within convolutional layers or blocks, based on which we gradually compute the outputs in a channel-wise manner.
The added orderings not only increase the depths and the learning capacities of the networks without any additional computation costs, but also eliminate the overlap singularities so that the networks are able to converge faster and perform better.
Experiments show that the networks based on our method achieve the state-of-the-art performances on CIFAR and ImageNet datasets.
\end{abstract}

\vspace{-0.2in}
\section{Introduction}
Deep neural networks have become the state-of-the-art systems for image recognition \cite{resnet,densenet,alexnet,fewshot,vggnet,GoogleNet,sort,znfnet}
as well as other vision tasks~\cite{deeplab,rich,fcnn,scalenet,fasterrcnn,deepcontour,hed}.
The architectures keep going deeper, \textit{e.g.}, from five convolutional layers \cite{alexnet} to $1001$ layers \cite{resnetv2}.
The benefit of deep architectures is their strong learning capacities because each new layer can potentially introduce more non-linearities and typically uses larger receptive fields~\cite{vggnet}.
In addition, adding certain types of layers (e.g.~\cite{resnetv2}) will not harm the performance theoretically since they can just learn identity mapping. This makes stacking up layers more appealing in the network designs.

\begin{figure}
    \centering
    \includegraphics[width=0.6\linewidth]{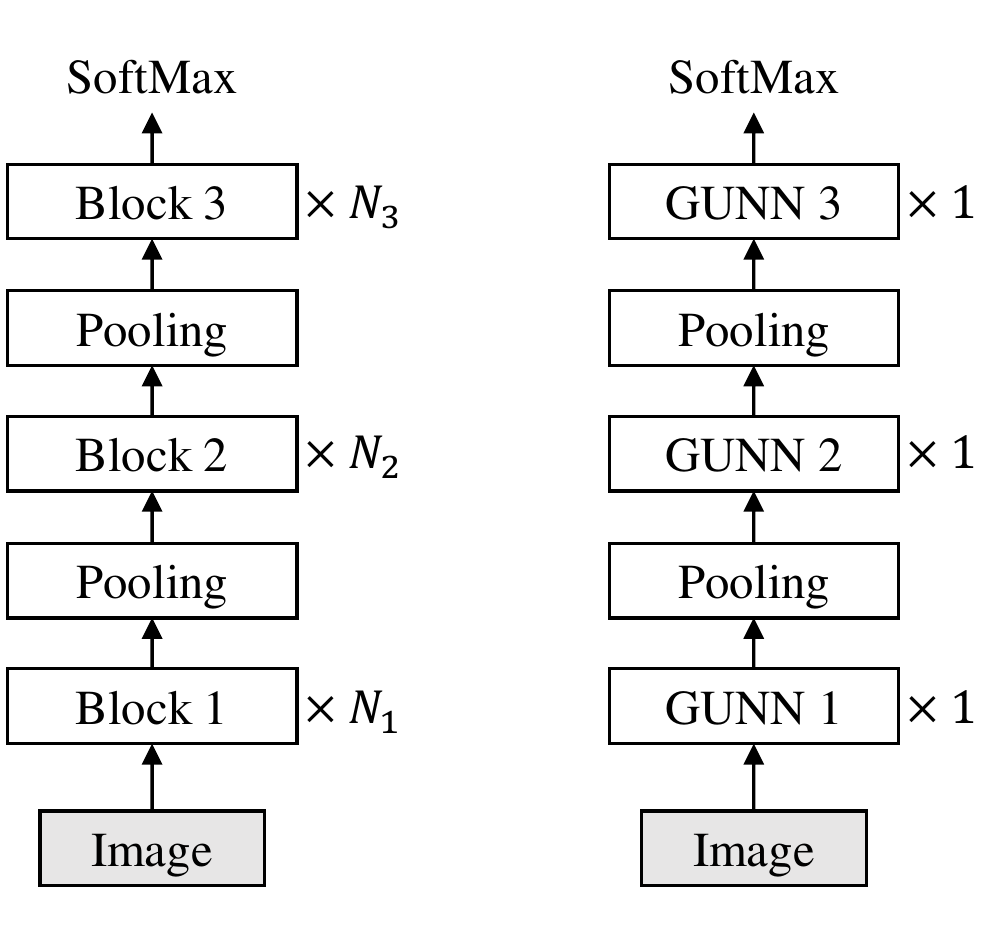}
    \caption{Comparing architecture designs based on cascading convolutional building blocks (left) and GUNN (right).
    Cascading-based architecture increases the depth by repeating the blocks. GUNN-based networks increases the depth by adding computation orderings to the channels of the building blocks.}
    \vspace{-0.2in}
    \label{fig:framework}
\end{figure}
\begin{figure*}
    \includegraphics[width=\linewidth]{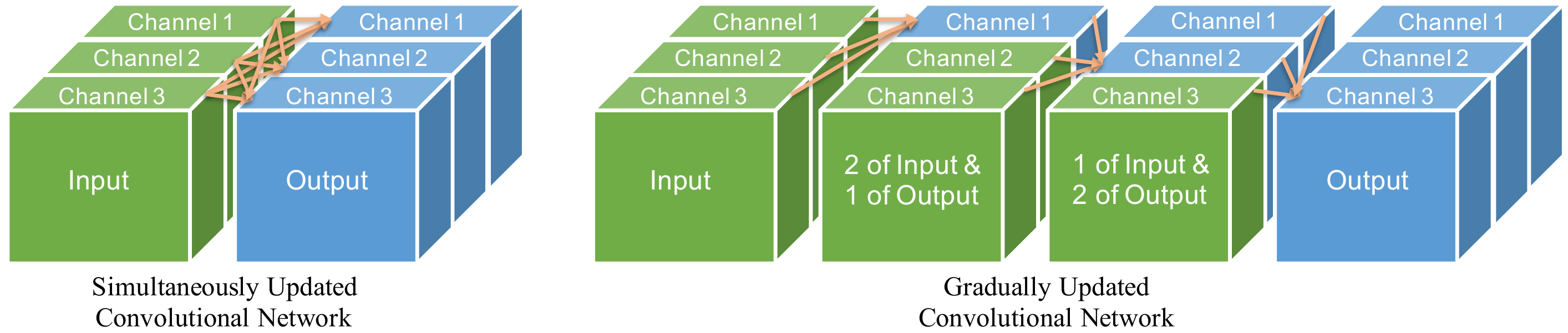}
    \vspace{-0.3in}
    \caption{Comparing Simultaneously Updated Convolutional Network and Gradually Updated Convolutional Network. Left is a traditional convolutional network with three channels in both the input and the output. Right is our proposed convolutional network which decomposes the original computation into three sequential channel-wise convolutional operations.
    In our proposed GUNN-based architectures, the updates are done by \emph{residual learning}~\cite{resnet}, which we do not show in this figure.}
    \vspace{-0.1in}
    \label{fig:new_arch}
\end{figure*}

Although deeper architectures usually lead to stronger learning capacities, cascading convolutional layers (\textit{e.g.} VGG~\cite{vggnet}) or blocks (\textit{e.g.} ResNet~\cite{resnet}) is not necessarily the only method to achieve this goal.
In this paper, we present a new way to increase the depth of the networks as an alternative to stacking up convolutional layers or blocks.
Figure~\ref{fig:new_arch} provides an illustration that compares our proposed convolutional network that gradually updates the feature representations against the traditional convolutional network that computes its output simultaneously.
By only adding an ordering to the channels without any additional computation, the later computed channels become deeper than the corresponding ones in the traditional convolutional network.
We refer to the neural networks with the proposed computation orderings on the channels as Gradually Updated Neural Networks (GUNN).
Figure~\ref{fig:framework} provides two examples of architecture designs based on cascading building blocks and GUNN.
Without repeating the building blocks, GUNN increases the depths of the networks as well as their learning capacities.

\begin{figure}
    \centering
    \includegraphics[width=0.9\linewidth]{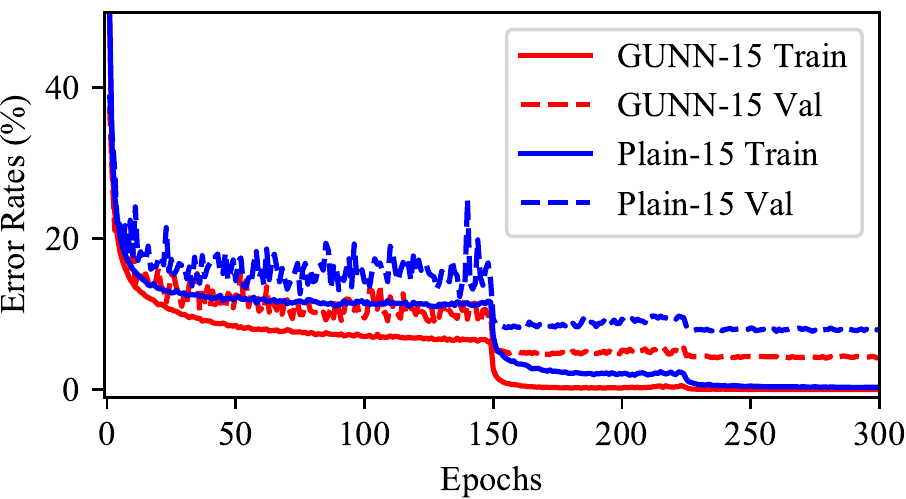}
    \vspace{-0.15in}
    \caption{Training dynamics on CIFAR-10 dataset.}
    \vspace{-0.2in}
    \label{fig:dynamic}
\end{figure}

It is clear that converting plain networks to GUNN increases the depths of the networks without any additional computations.
What is less obvious is that GUNN in fact eliminates the overlap singularities inherent in the loss landscapes of the cascading-based convolutional networks, which have been shown to adversely affect the training of deep neural networks as well as their performances~\cite{single,breaksym}.
Overlap singularity is when internal neurons collapse into each other, \textit{i.e.} they are unidentifiable by their activations.
It happens in the networks, increases the training difficulties and degrades the performances~\cite{breaksym}.
However, if a plain network is converted to GUNN, the added computation orderings will break the symmetry between the neurons.
We prove that the internal neurons in GUNN are impossible to collapse into each other.
As a result, the effective dimensionality can be kept during training and the model will be free from the degeneracy caused by collapsed neurons.
Reflected in the training dynamics and the performances, this means that converting to GUNN will make the plain networks \emph{easier to train} and \emph{perform better}.
Figure~\ref{fig:dynamic} compares the training dynamics of a 15-layer plain network on CIFAR-10 dataset~\cite{cifar} before and after converted to GUNN.

In this paper, we test our proposed GUNN on highly competitive benchmark datasets, \textit{i.e.} CIFAR~\cite{cifar} and ImageNet~\cite{ILSVRC15}.
Experimental results demonstrate that our proposed GUNN-based networks achieve the state-of-the-art performances compared with the previous cascading-based architectures.

\vspace{-0.25in}
\section{Related Work}
The research focuses of image recognition have moved from feature designs~\cite{hog,sift} to architecture designs~\cite{resnet,densenet,alexnet,overfeat,vggnet,GoogleNet,resnext,znfnet} due to the recent success of the deep neural networks.
Highway Networks~\cite{highway} proposed architectures that can be trained end-to-end with more than $100$ layers.
The main idea of Highway Networks is to use bypassing paths.
This idea was further investigated in ResNet~\cite{resnet}, which simplifies the bypassing paths by using only identity mappings.
As learning ultra-deep networks became possible, the depths of the models have increased tremendously.
ResNet with pre-activation~\cite{resnetv2} and ResNet with stochastic depth~\cite{stochastic} even managed to train neural networks with more than $1000$ layers.
FractalNet~\cite{fractalnet} argued that in addition to summation, concatenation also helps train a deep architecture.
More recently, ResNeXt~\cite{resnext} used group convolutions in ResNet and outperformed the original ResNet.
DenseNet~\cite{densenet} proposed an architecture with dense connections by feature concatenation.
Dual Path Net~\cite{dpn} finds a middle point between ResNet and DenseNet by concatenating them in two paths.
Unlike the above cascading-based methods, GUNN eliminates the overlap singularities caused by the architecture symmetry.
The detailed analyses can be found in Section 4.3.

\vspace{0.05in} 
Alternative to increasing the depth of the neural networks, another trend is to increase the widths of the networks.
GoogleNet~\cite{GoogleNet,GoogleNet2} proposed an \textit{Inception} module to concatenate feature maps produced by different filters.
Following ResNet~\cite{resnet}, the WideResNet~\cite{wideresnet} argued that compared with increasing the depth, increasing the width of the networks can be more effective in improving the performances.
Besides varying the width and the depth, there are also other design strategies for deep neural networks~\cite{hypercolumns,deepforest,ladder2,ladder1,dagcnn}.
Deeply-Supervised Nets~\cite{dsn} used auxiliary classifiers to provide direct supervisions for the internal layers.
Network in Network~\cite{nin} adds micro perceptrons to the convolutional layers.

\clearpage

\section{Model}\label{sec:model}
\subsection{Feature Update}
We consider a feature transformation $\mathcal{F}:\mathbb{R}^{m\times n}\rightarrow \mathbb{R}^{m\times n}$, where $n$ denotes the channel of the features and $m$ denotes the feature location on the $2$-D feature map.
For example, $\mathcal{F}$ can be a convolutional layer with $n$ channels for both the input and the output.
Let $\vect{x}\in\mathbb{R}^{m\times n}$ be the input and $\vect{y}\in\mathbb{R}^{m\times n}$ be the output, we have
\begin{equation}\label{eq:1}
    \vect{y} = \mathcal{F}(\vect{x})
\end{equation}
Suppose that $\mathcal{F}$ can be decomposed into channel-wise transformation $\mathcal{F}_c(\cdot)$ that are independent with eath other, then for any location $k$ and channel $c$ we have
\begin{equation}\label{eq:2}
    y_c^k = \mathcal{F}_c(\vect{x}^{r(k)})
\end{equation}
where $\vect{x}^{r(k)}$ denotes the receptive field of the location $k$ and $\mathcal{F}_c$ denotes the transformation on channel $c$.

Let $U_C$ denote a feature update on channel set $C$, \textit{i.e.},
\begin{equation}
\begin{aligned}
    U_C(\vect{x}):~~ & y_c^k=\mathcal{F}_c(\vect{x}^{r(k)}), \forall c\in C, k \\
    & y_c^k = x_c^k, \forall c\in\overline{C},k
\end{aligned}
\end{equation}
Then, $U_C=\mathcal{F}$ when $C=\{1,...,n\}$.

\subsection{Gradually Updated Neural Networks}
By defining the feature update $U_C$ on channel set $C$, the commonly used one-layer CNN is a special case of feature updates where every channel is updated simultaneously. However, we can also update the channels gradually.
For example, the proposed GUNN can be formulated by
\begin{equation}\label{eq:4}
    \begin{aligned}
    \text{GUNN}(\vect{x}) = (U_{c_l}\circ  U_{c_{(l-1)}} \circ ... \circ  U_{c_2}  \circ  U_{c_1})(\vect{x}) \\
    \text{where } \bigcup_{i=1}^{l} c_i = \{1,2,...,n\} \text{ and } c_i \cap c_j = \Phi,~\forall i\neq j
    \end{aligned}
\end{equation}
When $l=1$, GUNN is equivalent to $\mathcal{F}$.

Note that the number of parameters and computation of GUNN are the same as those of the corresponding $\mathcal{F}$ for any partitions $c_1, ..., c_l$ of $\{1,...,n\}$.
However, by decomposing $\mathcal{F}$ into channel-wise transformations and sequentially applying them, the later computed channels are deeper than the previous ones.
As a result, the depth of the network can be increased, as well as the network's learning capacity.

\subsection{Channel-wise Update by Residual Learning}

\begin{algorithm}
\SetKwInOut{Input}{Input}
\SetKwInOut{Output}{Output}
\caption{Back-propagation for GUNN}\label{alg:bp}
\Input{$U(\cdot) = (U_{c_l}\circ  U_{c_{(l-1)}} \circ ... \circ  U_{c_1})(\cdot)$, input $x$, \\
output $y=U(x)$, gradients $\partial L/\partial y$, \\
and parameters $\Theta$ for $U$.}
\Output{$\partial L/\partial \Theta$, $\partial L/\partial x$}
$\partial L/\partial x \leftarrow \partial L/\partial y$

\For{$i\leftarrow l$ \KwTo $1$}{
    $y_c\leftarrow x_c$, $\forall c\in c_i$

    $\partial L/\partial y, \partial L/\partial \Theta_{c_i} \leftarrow \text{BP}(y, \partial L/\partial x, U_{c_i}, \Theta_{c_i})$

    $(\partial L/\partial x)_c \leftarrow (\partial L/\partial y)_c$, $\forall c\in c_i$

    $(\partial L/\partial x)_c \leftarrow (\partial L/\partial x)_c + (\partial L/\partial y)_c$, $\forall c\not\in c_i$
}
\end{algorithm}

We consider the residual learning proposed by ResNet~\cite{resnet} in our model.
Specifically, we consider the channel-wise transformation $\mathcal{F}_c:\mathbb{R}^{m\times n}\rightarrow \mathbb{R}^{m\times 1}$ to be
\begin{equation}\label{eq:res}
  \mathcal{F}_c(x) = \mathcal{G}_c(x) + x_c
\end{equation}
where $\mathcal{G}_c$ is a convolutional neural network $\mathcal{G}_c:\mathbb{R}^{m\times n}\rightarrow \mathbb{R}^{m\times 1}$.
The motivation of expressing $\mathcal{F}$ in a residual learning manner is to reduce overlap singularities~\cite{breaksym}, which will be discussed in Section 4.

\vspace{-0.05in}
\subsection{Learning GUNN by Backpropagation}
Here we show the backpropagation algorithm for learning the parameters in GUNN that uses the same amount of computations and memory as in $\mathcal{F}$.
In Eq.~\ref{eq:4}, let the feature update $U_{c_i}$ be parameterized by $\Theta_{c_i}$.
Let $\text{BP}(x, \partial L/\partial y,f,\Theta)$ be the back-propagation algorithm for differentiable function $y=f(x;\Theta)$ with the loss $L$ and the parameters $\Theta$.
Algorithm~\ref{alg:bp} presents the back-propagation algorithm for GUNN. Since $U_{c_i}$ has the residual structures~\cite{resnet}, the last two steps can be merged into
\begin{equation}
    (\partial L/\partial x)_c \leftarrow (\partial L/\partial x)_c + (\partial L/\partial y)_c,~~\forall c\;
\end{equation}
which further simplifies the implementation.
It is easy to see that converting networks to GUNN-based does not increase the memory usage in feed-forwarding.
Given Algorithm~\ref{alg:bp}, converting networks to GUNN will not affect the memory in both the training and the evaluation.

\section{GUNN Eliminates Overlap Singularities}
Overlap singularities are inherent in the loss landscapes of some network architectures which are caused by the non-identifiability of subsets of the neurons.
They are identified and discussed in previous work~\cite{single,saddle,breaksym}, and are shown to be harmful for the performances of deep networks.
Intuitively, overlap singularities exist in architectures where the internal neurons collapse into each other.
As a result, the models are degenerate and the effective dimensionality is reduced.
\cite{breaksym} demonstrated through experiments that residual learning (see Eq.~\ref{eq:res}) helps to reduce the overlap singularities in deep networks, which partly explains the exceptional performances of ResNet~\cite{resnet} compared with plain networks.
In the following, we first use linear transformation as an example to demonstrate how GUNN-based networks break the overlap singularities.
Then, we generalize the results to ReLU DNN.
Finally, we compare GUNN with the previous state-of-the-art network architectures from the perspective of singularity elimination.

\subsection{Overlap Singularities in Linear Transformations}\label{sec:fc}
Consider a linear function $y=f(x): \mathbb{R}^n\rightarrow\mathbb{R}^n$ such that
\begin{equation}\label{eq:fc}
    y_i = \sum_{j=1}^n \omega_{i,j} x_j,~~\forall i\in\{1,..,n\}
\end{equation}
Suppose that there exists a pair of collapsed neurons $y_p$ and $y_q$ ($p<q$).
Then, for $\forall x$, $y_p=y_q$, and the equality holds after any number of gradient descents, \textit{i.e.} $\Delta y_p=\Delta y_q$.

Eq.~\ref{eq:fc} describes a plain network.
The solution for the existence of $y_p$ and $y_q$ is that
$\omega_{p,j}=\omega_{q,j},\forall j$.
This is the case that is mostly discussed previously, which happens in the networks and degrades the performances.

When we add the residual learning, Eq.~\ref{eq:fc} becomes
\begin{equation}\label{eq:fcres}
    y_i = x_i + \sum_{j=1}^n \omega_{i,j} x_j, ~~\forall i\in\{1,..,n\}
\end{equation}
Collapsed neurons require that $\omega_{p,p} + 1 = \omega_{q,p}$, $\omega_{q,q} + 1 = \omega_{p,q}$.
This will make the collapse of $y_p$ and $y_q$ very hard when $\omega$ is initialized from a normal distribution $\mathcal{N}(0, \sqrt{2/n})$ as in ResNet, but still possible.

Next, we convert Eq.~\ref{eq:fcres}  to GUNN, \textit{i.e.,}
\begin{equation}\label{eq:9}
  y_i = x_i + \sum_{j=1}^{i-1}\omega_{i,j}y_j + \sum_{j=i}^{n}\omega_{i,j}x_j, ~~\forall i\in\{1,..,n\}
\end{equation}
Suppose that $y_p$ and $y_q$ ($p<q$) collapse.
Consider $\Delta y$, the value difference at $x$ after one step of gradient descent on $\omega$ with input $x$, $\partial L/\partial y$ and learning rate $\epsilon$. When $\epsilon\rightarrow 0$,
\begin{equation}\label{eq:10}
    \Delta y_i = \epsilon\dfrac{\partial L}{\partial y_i}(\sum_{j=1}^{i-1}y_j^2 + \sum_{j=i}^{n}x_j^2) + \sum_{j=1}^{i-1} \omega_{i,j}\Delta y_j
\end{equation}
As $\Delta y_p=\Delta y_q, \forall x$, we have $\omega_{q,j} = 0,~~\forall j: p< j < q$.
But this condition will be broken in the next update;
thus, $q=p+1$.
Then, we derive that $y_p=y_q=0$.
But these will also be broken in the next step of gradient descent optimization.
Hence, $y_p$ and $y_q$ cannot collapse into each other.
The complete proof can be found in the appendix.

\subsection{Overlap Singularities in ReLU DNN}
In practice, architectures are usually composed of several linear layers and non-linearity layers.
Analyzing all the possible architectures is beyond our scope.
Here, we discuss the commonly used ReLU DNN, in which only linear transformations and ReLUs are used by simple layer cascading.

Following the notations in \S\ref{sec:model}, we use $y=\mathcal{G}(x)+x$, in which $\mathcal{G}(x)$ is a ReLU DNN.
Note that $\mathcal{G}$ is continuous piecewise linear (PWL) function~\cite{reludnn}, which means that there exists a finite set of polyhedra whose union is $\mathbb{R}^n$, and $\mathcal{G}$ is affine linear over each polyhedron.

Suppose that we convert $\mathcal{G}(x) + x$ to GUNN and there exists a pair of collapsed neurons $y_p$ and $y_q$ ($p<q$).
Then, the set of polyhedra for $y_p$ is the same as for $y_q$.
Let $\mathbb{P}$ be a polyhedron for $y_p$ and $y_q$ defined above.
Then, $\forall x, \mathbb{P}, i$,
\begin{equation}
  y_i = x_i + \sum_{j=1}^{i-1}\omega_{i,j}(\mathbb{P})y_j + \sum_{j=i}^{n}\omega_{i,j}(\mathbb{P})x_j
\end{equation}
where $\omega(\mathbb{P})$ denotes the parameters for polyhedron $\mathbb{P}$.
Note that on each $\mathbb{P}$, $y$ is a function of $x$ in the form of Eq.~\ref{eq:9}; hence, $y_p$ and $y_q$ cannot collapse into each other.
Since the union of all polyhedra is $\mathbb{R}^n$, we conclude that GUNN eliminates the overlap singularities in ReLU DNN.

\subsection{Discussions and Comparisons}
The previous two subsections consider the GUNN conversion where $|c_i|=1, \forall i$ (see Eq.~\ref{eq:4}).
But this will slow down the computation on GPU due to the data dependency.
Without specialized hardware or library support, we decide to increase $|c_i|$ to $>10$.
The resulted models run at the speed between ResNeXt~\cite{resnext} and DenseNet~\cite{densenet}.
But this change introduces singularities into the channels from the same set $c_i$.
Then, the residual learning helps GUNN to reduce the singularities within the same set $c_i$ since we initialize the parameters from a normal distribution $\mathcal{N}(0, \sqrt{2/n})$.
We will compare the results of GUNN with and without residual learning in the experiments.

We compare GUNN with the state-of-the-art architectures from the perspective of overlap singularities.
ResNet~\cite{resnet} and its variants use residual learning, which reduces but cannot eliminate the singularities.
ResNeXt~\cite{resnext} uses group convolutions to break the symmetry between groups, which further helps to avoid neuron collapses.
DenseNet~\cite{densenet} concatenates the outputs of layers as the input to the next layer.
DenseNet and GUNN both create dense connections, while DenseNet reuses the outputs by concatenating and GUNN by adding them back to the inputs.
But the channels within the same layer of DenseNet are still possible to collapse into each other since they are symmetric.
In contrast, adding back makes residual learning possible in GUNN.
This makes residual learning indispensable in GUNN-based networks.

\section{Network Architectures}
In this section, we will present the details of our architectures for the CIFAR~\cite{cifar} and ImageNet~\cite{ILSVRC15} datasets.

\subsection{Simultaneously Updated Neural Networks and Gradually Updated Neural Networks}
Since the proposed GUNN is a method for increasing the depths of the convolutional networks, specifying the architectures to be converted is equivalent to specifying the GUNN-based architectures.
The architectures before conversion, the Simultaneously Updated Neural Networks (SUNN), become natural baselines for our proposed GUNN networks.
We first study what baseline architectures can be converted.

There are two assumptions about the feature transformation $\mathcal{F}$ (see Eq.~\ref{eq:1}): (1) the input and the output sizes are the same, and (2) $\mathcal{F}$ is channel-wise decomposable.
To satisfy the first assumption, we will first use a convolutional layer with Batch Normalization~\cite{batchnorm} and ReLU~\cite{relu} to transform the feature space to a new space where the number of the channels is wanted.
To satisfy the second assumption, instead of directly specifying the transform $\mathcal{F}$, we focus on designing $\mathcal{F}_{c_i}$, where $c_i$ is a subset of the channels (see Eq.~\ref{eq:4}).
To be consistent with the term \emph{update} used in GUNN and SUNN, we refer to $\mathcal{F}_{c_i}$ as the \emph{update units} for channels $c_i$.

\paragraph{Bottleneck Update Units}\label{sec:buu}
\begin{figure}[!htp]
    \centering
    \includegraphics[width=0.6\linewidth]{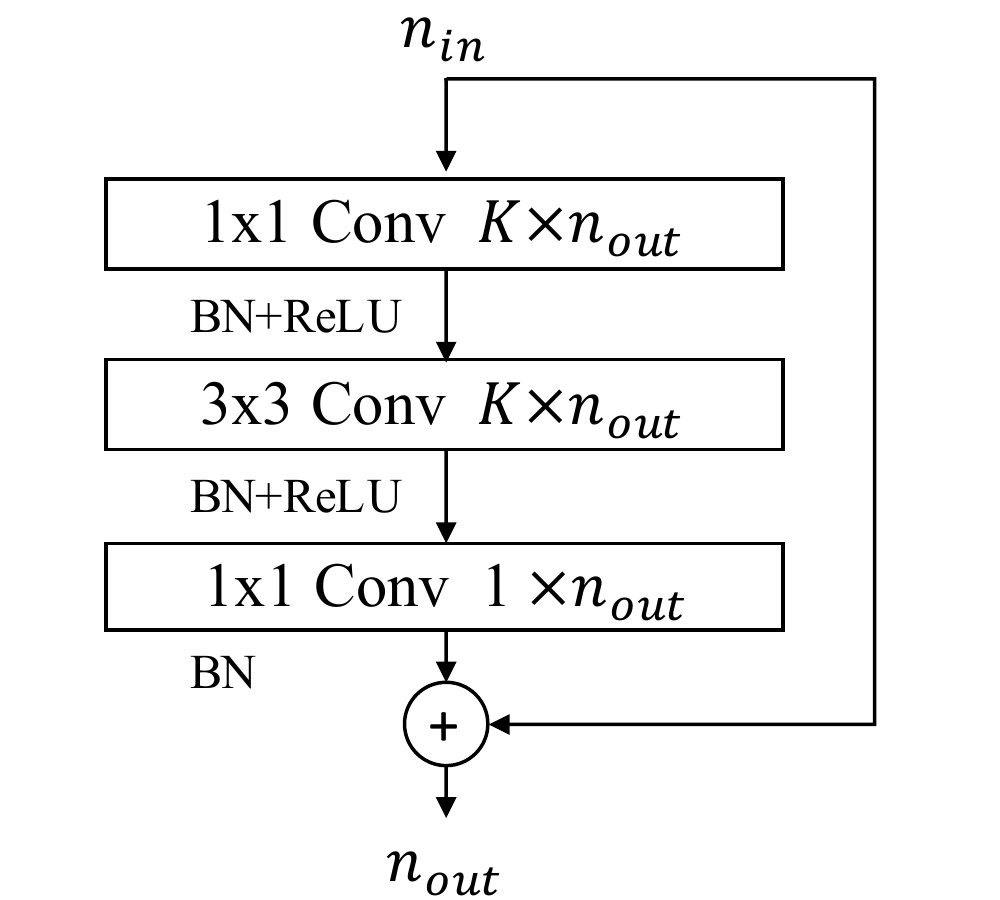}
    \caption{Bottleneck Update Units for both SUNN and GUNN.}
    \label{fig:uu}
\end{figure}
In the architectures proposed in this paper, we adopt bottleneck neural networks as shown in Figure~\ref{fig:uu} for the update units for both the SUNN and GUNN.
Suppose that the update unit maps the input features of channel size $n_{\text{in}}$ to the output features of size $n_{\text{out}}$.
Each unit contains three convolutional layers.
The first convolutional layer transforms the input features to $K\times n_{\text{out}}$ using a $1\times1$ convolutional layer.
The second convolutional layer is of kernel size $3\times3$, stride $1$, and padding $1$, outputting the features of size $K\times n_{\text{out}}$.
The third layer computes the features of size $n_{\text{out}}$ using a $1\times1$ convolutional layer.
The output is then added back to the input, following the residual architecture proposed in ResNet~\cite{resnet}.
We add batch normalization layer~\cite{batchnorm} and ReLU layer~\cite{relu} after the first and the second convolutional layers, while only adding batch normalization layer after the third layer.
Stacking up $M$ update units also generates a new one.
In total, we have two hyperparameters for designing an update unit: the expansion rate $K$ and the number of the $3$-layer update units $M$.

\vspace{-0.1in}
\paragraph{One Resolution, One Representation}

Our architectures will have only one representation at one resolution besides the pooling layers and the convolutional layers that initialize the needed numbers of channels.
Take the architecture in Table~\ref{tab:1} as an example.
There are two processes for each resolution.
The first one is the transition process, which computes the initial features with the dimensions of the next resolution, then down samples it to $1/4$ using a $2\times2$ average pooling.
A convolutional operation is needed here because $\mathcal{F}$ is assumed to have the same input and output sizes.
The next process is using GUNN to update this feature space gradually.
Each channel will only be updated once, and all channels will be updated after this process.
Unlike most of the previous networks, after this two processes, the feature transformations at this resolution are complete.
There will be no more convolutional layers or blocks following this feature representation, \textit{i.e.}, \textit{one resolution, one representation}.
Then, the network will compute the initial features for the next resolution, or compute the final vector representation of the entire image by a global average pooling.
By designing networks in this way, SUNN networks usually have about $20$ layers before converting to GUNN-based networks.

\begin{table}[!ht]
  \small
  \centering
  \begin{tabular}{l|c|c|c}
    \toprule
    Stage & Output & WideResNet-$28$-$10$ & GUNN-$15$ \\
    \midrule
    Conv1 & $32\times32$  & $ \begin{bmatrix} 3\times3, 16 \end{bmatrix}$       & $ \begin{bmatrix} 3\times3, 64 \\
    1\times1, 240 \end{bmatrix}$ \\
    \midrule
    Conv2$^*$ & $32\times32$  & $ \begin{bmatrix} 3\times3, 160 \\
    3\times3, 160 \end{bmatrix} \times4$       & $ \begin{bmatrix} 1\times1, \times2 \\
    3\times3, \times2 \\
    1\times1, \times1 \end{bmatrix}$ \\
    \midrule
    Trans1 & $16\times16$ &  -- & $ \begin{bmatrix} 1\times1, 300 \\
    \text{AvgPool} \end{bmatrix}$ \\
    \midrule
    Conv3$^*$ & $16\times16$  & $ \begin{bmatrix} 3\times3, 320 \\
    3\times3, 320 \end{bmatrix} \times4$       & $ \begin{bmatrix} 1\times1, \times2 \\
    3\times3, \times2 \\
    1\times1, \times1 \end{bmatrix}$ \\
    \midrule
    Trans2 & $8\times8$ &  -- & $ \begin{bmatrix} 1\times1, 360 \\
    \text{AvgPool} \end{bmatrix}$ \\
    \midrule
    Conv4$^*$ & $8\times8$  & $ \begin{bmatrix} 3\times3, 640 \\
    3\times3, 640 \end{bmatrix} \times4$       & $ \begin{bmatrix} 1\times1, \times2 \\
    3\times3, \times2 \\
    1\times1, \times1 \end{bmatrix}$ \\
    \midrule
    Trans3 & $1\times1$ & $ \begin{bmatrix} \text{GAPool} \\ \text{fc, softmax} \end{bmatrix}$ & $ \begin{bmatrix} 1\times1, 360 \\
    \text{GAPool} \\ \text{fc, softmax} \end{bmatrix}$ \\
    \midrule
    \multicolumn{2}{l|}{GPU Memory} & $4.903$GB@$64$ & $2.485$GB@$64$ \\
    \midrule
    \multicolumn{2}{l|}{\# Params} & $36.5$M  & $1.6$M \\
    \midrule
    \multicolumn{2}{l|}{Error (C10/C100)} & $4.17$ / $20.50$ & $4.15$ / $20.45$ \\
    \bottomrule
  \end{tabular}
  \vspace{0.1in}
  \caption{Architecture comparison between WideResNet-28-10~\cite{wideresnet} and GUNN-15 for CIFAR. (Left) WideResNet-28-10. (Right) GUNN-15.
  GUNN achieves comparable accuracies on CIFAR10/100 while using a smaller number of parameters and consuming less GPU memory during training.
  In GUNN-15, the convolution stages with stars are computed using GUNN while others are not.}\label{tab:1}
\end{table}

\vspace{-0.1in}
\paragraph{Channel Partitions}
With the clearly defined update units, we can easily build SUNN and GUNN layers by using the units to update the representations following Eq.~\ref{eq:4}.
The hyperparameters for the SUNN/GUNN layer are the number of the channels $N$ and the partition over those channels.
In our proposed architectures, we evenly partition the channels into $P$ segments.
Then, we can use $N$ and $P$ to represent the configuration of a layer.
Together with the hyperparameters in the update units, we have four hyperparameters to tune for one SUNN/GUNN layer, \textit{i.e.} \{$N,P,K,M$\}.

\subsection{Architectures for CIFAR}
We have implemented two neural networks based on GUNN to compete with the previous state-of-the-art methods on CIFAR datasets, \textit{i.e.}, GUNN-$15$ and GUNN-$24$.
Table~\ref{tab:1} shows the big picture of GUNN-$15$.
Here, we present the details of the hyperparameter settings for GUNN-$15$ and GUNN-$24$.
For GUNN-$15$, we have three GUNN layers, Conv$2$, Conv$3$ and Conv$4$. The configuration for Conv$2$ is $\{ N=240, P=20, K=2, M=1 \}$, the configuration for Conv$3$ is $\{ N=300, P=25, K=2, M=1 \}$, and the configuration for Conv$4$ is $\{ N=360, P=30, K=2, M=1 \}$.
For GUNN-$24$, based on GUNN-$15$, we change the number of output channels of Conv$1$ to $720$, Trans$1$ to $900$, Trans$2$ to $1080$, and Trans$3$ to $1080$.
The hyperparameters are $\{ N=720, P=20, K=3, M=2 \}$ for Conv$2$, $\{ N=900, P=25, K=3, M=2 \}$ for Conv$3$, and $\{ N=1080, P=30, K=3, M=2 \}$ for Conv$3$.
The number of parameters of GUNN-$15$ is $1585746$ for CIFAR-10 and $1618236$ for CIFAR-100. The number of parameters of GUNN-$24$ is $29534106$ for CIFAR-10 and $29631396$ for CIFAR-100.
The GUNN-$15$ is aimed to compete with the methods published in an early stage by using a much smaller model, while GUNN-$24$ is targeted at comparing with ResNeXt~\cite{resnext} and DenseNet~\cite{densenet} to get the state-of-the-art performance.

\subsection{Architectures for ImageNet}
\begin{table}[!ht]
  \setlength{\tabcolsep}{2pt}
  \small
  \centering
  \begin{tabular}{l|c|c|c}
    \toprule
    Stage & Output & ResNet-$152$ & GUNN-$18$ \\
    \midrule
    Conv1 & $112\times112$  & $ \begin{bmatrix} 7\times7, 64, 2 \\
    3\times3~\text{MaxPool}, 2 \end{bmatrix}$       & $ \begin{bmatrix} 7\times7, 64, 2\\
    3\times3~\text{MaxPool}, 2 \\
    1\times1, 400 \end{bmatrix}$ \\
    \midrule
    Conv2$^*$ & $112\times112$  & $ \begin{bmatrix} 1\times1, 64~~ \\
    3\times3, 64~~ \\
    1\times1, 256 \end{bmatrix} \times3$       & $ \begin{bmatrix} 1\times1, \times2 \\
    3\times3, \times2 \\
    1\times1, \times1 \end{bmatrix}$ \\
    \midrule
    Trans1 & $56\times56$ &  -- & $ \begin{bmatrix} 1\times1, 800 \\
    \text{AvgPool} \end{bmatrix}$ \\
    \midrule
    Conv3$^*$ & $56\times56$  & $ \begin{bmatrix} 1\times1, 128 \\
    3\times3, 128 \\
    1\times1, 512 \end{bmatrix} \times8$       & $ \begin{bmatrix} 1\times1, \times2 \\
    3\times3, \times2 \\
    1\times1, \times1 \end{bmatrix}$ \\
    \midrule
    Trans2 & $28\times28$ &  -- & $ \begin{bmatrix} 1\times1, 1600 \\
    \text{AvgPool} \end{bmatrix}$ \\
    \midrule
    Conv4$^*$ & $28\times28$  &  $ \begin{bmatrix} 1\times1, 256~~ \\
    3\times3, 256~~ \\
    1\times1, 1024 \end{bmatrix} \times36$      & $ \begin{bmatrix} 1\times1, \times2 \\
    3\times3, \times2 \\
    1\times1, \times1 \end{bmatrix}$ \\
    \midrule
    Trans3 & $14\times14$ &  -- & $ \begin{bmatrix} 1\times1, 2000 \\
    \text{AvgPool} \end{bmatrix}$ \\
    \midrule
    Conv5$^*$ & $14\times14$  &  $ \begin{bmatrix} 1\times1, 512~~ \\
    3\times3, 512~~ \\
    1\times1, 2048 \end{bmatrix} \times3$      & $ \begin{bmatrix} 1\times1, \times2 \\
    3\times3, \times2 \\
    1\times1, \times1 \end{bmatrix}$ \\
    \midrule
    Trans4 & $\times1$ & $ \begin{bmatrix} \text{GAPool} \\ \text{fc, softmax} \end{bmatrix}$ & $ \begin{bmatrix}
    \text{GAPool} \\ \text{fc, softmax} \end{bmatrix}$ \\
    \midrule
    \multicolumn{2}{l|}{\# Params} & $60.2$M  & $28.9$M \\
    \midrule
    \multicolumn{2}{l|}{Error (Top-1/5)} & $22.2$ / $6.2$ & $21.65$ / $5.87$ \\
    \bottomrule
  \end{tabular}
  \vspace{0.1in}
  \caption{Architecture comparison between ResNet~\cite{resnet} and GUNN-18 for ImageNet-152. (Left) ResNet-152. (Right) GUNN-18.
  GUNN achieves better accuracies on ImageNet while using a smaller number of parameters.}\label{tab:2}
\end{table}

We implement a neural network GUNN-$18$ to compete with the state-of-the-art neural networks on ImageNet with a similar number of parameters.
Table~\ref{tab:2} shows the big picture of the neural network architecture of GUNN-$18$.
Here, we present the detailed hyperparameters for the GUNN layers in GUNN-$18$.
The GUNN layers include Conv2, Conv3, Conv4 and Conv5.
The hyperparameters are $\{ N=400, P=10, K=2, M=1 \}$ for Conv2, $\{ N=800, P=20, K=2, M=1 \}$ for Conv3, $\{ N=1600, P=40, K=2, M=1 \}$ for Conv4 and $\{ N=2000, P=50, K=2, M=1 \}$ for Conv5. The number of parameters is $28909736$.
The GUNN-$18$ is targeted at competing with the previous state-of-the-art methods that have similar numbers of parameters, \textit{e.g.}, ResNet-$50$~\cite{resnext},   ResNeXt-$50$~\cite{resnext} and DenseNet-$264$~\cite{densenet}.

We also implement a wider GUNN-based neural networks \emph{Wide-GUNN-18} for better capacities.
The hyperparameters are $\{ N=1200, P=30, K=2, M=1 \}$ for Conv2, $\{ N=1600, P=40, K=2, M=1 \}$ for Conv3, $\{ N=2000, P=50, K=2, M=1 \}$ for Conv4 and $\{ N=2000, P=50, K=2, M=1 \}$ for Conv5.
The number of parameters is $45624936$.
The Wide-GUNN-$18$ is targeted at competing with ResNet-$101$, ResNext-$101$, DPN~\cite{dpn} and SENet~\cite{senet}.

\begin{table*}
  \setlength{\tabcolsep}{10pt}
  \small
  \centering
  \begin{tabular}{l|c|c|c|c}
    \toprule
    \multicolumn{3}{c|}{Method} & C10 & C100 \\
    \midrule
    \multicolumn{3}{c|}{Network in Network~\cite{nin}} & $8.81$ & -- \\
    \multicolumn{3}{c|}{All-CNN~\cite{allconv}} & $7.25$ & $33.71$ \\
    \multicolumn{3}{c|}{Deeply Supervised Network~\cite{dsn}} & $7.97$ & $34.57$\\
    \multicolumn{3}{c|}{Highway Network~\cite{highway}} & $7.72$ & $32.39$\\
    \midrule
    & \# layers & \# params & \\
    \midrule
    ResNet~\cite{resnet,stochastic} & $110$ & $1.7$M & $6.41$ & $27.22$\\
    FractalNet~\cite{fractalnet} & $21$ & $38.6$M & $5.22$ & $23.30$ \\
    Stochastic Depth~\cite{stochastic} & $1202$ &  $10.2$M & $4.91$ & $24.58$ \\
    ResNet with pre-act~\cite{resnetv2} & $1001$ & $10.2$M & $4.62$ & $22.71$ \\
    WideResNet-$28$-$10$~\cite{wideresnet} & $28$ & $36.5$M & $4.17$ & $20.50$\\
    WideResNet-$40$-$10$~\cite{wideresnet} & $40$ & $55.8$M & $3.80$ & $18.30$\\
    \midrule
    ResNeXt~\cite{resnext} & $29$ & $68.1$M & $3.58$ & $17.31$ \\
    DenseNet~\cite{densenet} & $190$ & $25.6$M & $3.46$ & $17.18$ \\
    Snapshot Ensemble~\cite{snapshot} & -- & $163.2$M & $3.44$ & $17.41$ \\
    \midrule
    GUNN-15 & $15$ & $1.6$M & $4.15$ & $20.45$ \\
    GUNN-24 & $24$ & $29.6$M & $\mathbf{3.21}$ & $\mathbf{16.69}$ \\
    GUNN-24 Ensemble & $24\times 6$ & $177.6$M & $\mathbf{3.02}$ & $\mathbf{15.61}$ \\
    \bottomrule
  \end{tabular}
  \vspace{0.1in}
  \caption{Classification errors (\%) on the CIFAR-10/100 test set. All methods are with data augmentation.
  The third group shows the most recent state-of-the-art methods.
  The performances of GUNN are presented in the fourth group.
  A very small model GUNN-$15$ outperforms all the methods in the second group except WideResNet-$40$-$10$.
  A relatively bigger model GUNN-$24$ surpasses all the competing methods.
  GUNN-$24$ becomes more powerful with ensemble~\cite{snapshot}.}\label{tab:c_res}
\end{table*}

\section{Experiments}
In this section, we demonstrate the effectiveness of the proposed GUNN on
several benchmark datasets.

\subsection{Benchmark Datasets}
\paragraph{CIFAR} CIFAR~\cite{cifar} has two color image datasets: CIFAR-10 (C10) and CIFAR-100 (C100).
Both datasets consist of natural images with the size of $32\times 32$ pixels.
The CIFAR-10 dataset has $10$ categories, while the CIFAR-100 dataset has $100$ categories.
For both of the datasets, the training and test set contain $50,000$ and $10,000$ images, respectively.
To fairly compare our method with the state-of-the-arts~\cite{resnet,densenet,stochastic,fractalnet,dsn,nin,fitnet,allconv,highway,resnext}, we use the same training and testing strategies, as well as the data processing methods.
Specifically, we adopt a commonly used data augmentation scheme, \textit{i.e.,} mirroring and shifting, for these two datasets.
We use channel means and standard derivations to normalize the images for data pre-processing.

\vspace{-0.15in}
\paragraph{ImageNet} The ImageNet dataset~\cite{ILSVRC15} contains about $1.28$ million color images for training and $50,000$ for validation.
The dataset has $1000$ categories.
We adopt the same data augmentation methods as in the state-of-the-art architectures~\cite{resnet,resnetv2,densenet,resnext} for training.
For testing, we use single-crop at the size of $224\times224$.
Following the state-of-the-arts~\cite{resnet,resnetv2,densenet,resnext}, we report the validation error rates.

\vspace{-0.03in}
\subsection{Training Details}
We train all of our networks using stochastic gradient descents.
On CIFAR-10/100~\cite{cifar}, the initial learning rate is set to $0.1$, the weight decay is set to $1e^{-4}$, and the momentum is set to $0.9$ without dampening.
We train the models for $300$ epochs.
The learning rate is divided by $10$ at $150$th epoch and $225$th epoch.
We set the batch size to $64$, following~\cite{densenet}.
All the results reported for CIFAR, regardless of the detailed configurations, were trained using $4$ NVIDIA Titan X GPUs with the data parallelism.
On ImageNet~\cite{ILSVRC15}, the learning rate is also set to $0.1$ initially, and decreases following the schedule in DenseNet~\cite{densenet}.
The batch size is set to $256$.
The network parameters are also initialized following~\cite{resnet}.
We use $8$ Tesla V100 GPUs with the data parallelism to get the reported results.
Our results are directly comparable with ResNet, WideResNet, ResNeXt and DenseNet.

\subsection{Results on CIFAR}
We train two models GUNN-$15$ and GUNN-$24$ for the CIFAR-10/100 dataset.
Table~\ref{tab:c_res} shows the comparisons between our method and the previous state-of-the-art methods. Our method GUNN achieves the best results in the test of both the single model and the ensemble test.
Here, we use Snapshot Ensemble~\cite{snapshot}.

\begin{table*}
  \setlength{\tabcolsep}{10pt}
  \small
  \centering
  \begin{tabular}{l|c|c|c|c}
    \toprule
    Method & \# layers & \# params & top-$1$ & top-$5$ \\
    \midrule
    VGG-$16$~\cite{vggnet} & $16$ & $138$M & $28.5$ & $9.9$ \\
    ResNet-$50$~\cite{resnet} & $50$ & $25.6$M & $24.0$ & $7.0$ \\
    ResNeXt-$50$~\cite{resnext} & $50$ & $25.0$M & $22.2$ & $6.0$\\
    DenseNet-$264$~\cite{densenet} & $264$ & $33.3$M & $22.15$ & $6.12$\\
    SUNN-$18^*$ & $18$ & $28.9$M & $26.16$ & $8.48$ \\
    GUNN-$18^*$ & $18$ & $28.9$M & $\mathbf{21.65}$ & $\mathbf{5.87}$ \\
    \midrule
    ResNet-$101$~\cite{resnet} & $101$ & $44.5$M & $22.0$ & $6.0$ \\
    ResNeXt-$101$~\cite{resnext} & $101$ & $44.1$M & $21.2$ & $5.6$ \\
    DPN-$98$~\cite{dpn} & $98$ & $37.7$M & $20.73$ & $5.37$ \\
    SE-ResNeXt-$101$~\cite{senet} & $101$ & $49.0$M & $20.70$ & $\mathbf{5.01}$ \\
    Wide GUNN-$18^*$ & $18$ & $45.6$M & $\mathbf{20.59}$ & $5.52$ \\
    \bottomrule
  \end{tabular}
  \vspace{0.1in}
  \caption{Single-crop classification errors (\%) on the ImageNet validation set. The test size of all the methods is $224\times224$.
  Ours: $*$.}\label{tab:i_res}
\end{table*}

\vspace{-0.1in}
\paragraph{Baseline Methods}
Here we present the details of baseline methods in Table~\ref{tab:c_res}.
The performances of ResNet~\cite{resnet} are reported in Stochastic Depth~\cite{stochastic} for both C10 and C100.
The WideResNet~\cite{wideresnet} WRN-$40$-$10$ is reported in their official code repository on GitHub.
The ResNeXt in the third group is of configuration $16\times64$d, which has the best result reported in the paper~\cite{resnext}.
The DenseNet is of configuration DenseNet-BC ($k=40$), which achieves the best performances on CIFAR-10/100.
The Snapshot Ensemble~\cite{snapshot} uses $6$ DenseNet-$100$ to ensemble during inference.
We do not compare with methods that use more data augmentation (\textit{e.g.} \cite{mixup}) or stronger regularizations (\textit{e.g.} \cite{shake}) for the fairness of comparison.


\vspace{-0.1in}
\paragraph{Ablation Study}
For ablation study, we compare GUNN with SUNN, \textit{i.e.}, the networks before the conversion.
Table~\ref{tab:c_ablation} shows the comparison results, which demonstrate the effectiveness of GUNN.
We also compare the performances of GUNN with and without residual learning.

\begin{table}
  \setlength{\tabcolsep}{8pt}
  \small
  \centering
  \begin{tabular}{l|c|c|c|c}
    \toprule
    Method & \# layers & \# params & C10 & C100\\
    \midrule
    GUNN-15-NoRes & $15$ & $1.6$M & $4.45$ & $21.15$ \\
    GUNN-15       & $15$ & $1.6$M & $4.15$ & $20.45$ \\
    \midrule
    SUNN-15 & $15$ & $1.6$M & $5.64$ & $23.75$ \\
    GUNN-15 & $15$ & $1.6$M & $4.15$ & $20.45$ \\
    \midrule
    SUNN-24 & $24$ & $29.6$M & $3.88$ & $19.60$ \\
    GUNN-24 & $24$ & $29.6$M & $3.21$ & $16.69$ \\
    \bottomrule
  \end{tabular}
  \vspace{0.1in}
  \caption{Ablation study on residual learning and SUNN.}\label{tab:c_ablation}
\end{table}

\subsection{Results on ImageNet}
We evaluate the GUNN on the ImageNet classification task, and compare our performances with the state-of-the-art methods.
These methods include VGGNet~\cite{vggnet}, ResNet~\cite{resnet}, ResNeXt~\cite{resnext}, DenseNet~\cite{densenet}, DPN~\cite{dpn} and SENet~\cite{senet}.
The comparisons are shown in Table~\ref{tab:i_res}.
The results of ours, ResNeXt, and DenseNet are directly comparable as these methods use the same framework for training and testing networks.
Table~\ref{tab:i_res} groups the methods by their numbers of parameters, except VGGNet which has $1.38\times10^{8}$ parameters.

The results presented in Table~\ref{tab:i_res} demonstrate that with the similar number of parameters, GUNN can achieve comparable performances with the previous state-of-the-art methods.
For GUNN-$18$, we also conduct an ablation experiment by comparing the corresponding SUNN with GUNN of the same configuration.
Consistent with the experimental results on the CIFAR-10/100 dataset, the proposed GUNN improves the accuracy on ImageNet dataset.

\vspace{-0.06in}
\section{Conclusions}
In this paper, we propose Gradually Updated Neural Network (GUNN), a novel, simple yet effective method to increase the depths of neural networks as an alternative to cascading layers.
GUNN is based on Convolutional Neural Networks (CNNs), but differs from CNNs in the way of computing outputs.
The outputs of GUNN are computed gradually rather than simultaneously as in CNNs in order to increase the depth.
Essentially, GUNN assumes the input and the output are of the same size and adds a computation ordering to the channels.
The added ordering increases the receptive fields and non-linearities of the later computed channels.
Moreover, it eliminates the overlap singularities inherent in the traditional convolutional networks.
We test GUNN on the task of image recognition.
The evaluations are done in three highly competitive benchmarks, CIFAR-10, CIFAR-100 and ImageNet.
The experimental results demonstrate the effectiveness of the proposed GUNN on image recognition.
In the future, since the proposed GUNN can be used to replace CNNs in other neural networks, we will study the applications of GUNN in other visual tasks, such as object detection and semantic segmentation.


\bibliography{example_paper}
\bibliographystyle{icml2018}

\end{document}